\documentclass[conference, letterpaper]{ieeeconf}
\IEEEoverridecommandlockouts

\usepackage[accsupp]{axessibility}
\usepackage{booktabs}
\usepackage{colortbl}
\usepackage{graphicx}
\usepackage[hidelinks]{hyperref}
\usepackage{multirow}
\usepackage{siunitx}
\usepackage[caption=false]{subfig}
\usepackage{xcolor}
\usepackage{xspace}
\usepackage{glossaries}

\usepackage{amssymb}
\usepackage{pifont}
\newcommand{\cmark}{\ding{51}}%
\usepackage{tikz}
\usetikzlibrary{positioning}

\makeatletter
\DeclareRobustCommand\onedot{\futurelet\@let@token\@onedot}
\def\@onedot{\ifx\@let@token.\else.\null\fi\xspace}

\def\eg{\emph{e.g}\onedot} 
\def\ie{\emph{i.e}\onedot}

\def\etal{\emph{et~al}\onedot}
\makeatother

\newacronym{fov}{FOV}{Field Of View}
\newacronym{imu}{IMU}{Inertial Measurement Unit}
\newacronym{rtk}{RTK}{Real-Time Kinematics}
\newacronym{ptp}{PTP}{Precision Time Protocol}
\newacronym{fmcw}{FMCW}{Frequency Modulated Continuous Wave}

\newcommand\copyrighttext{\footnotesize \textcopyright~2024 IEEE. Personal use of this material is permitted. Permission from IEEE must be obtained for all other uses, in any current or future media, including reprinting/republishing this material for advertising or promotional purposes, creating new collective works, for resale or redistribution to servers or lists, or reuse of any copyrighted component of this work in other works.
}

\newcommand\copyrightnotice{%
    \begin{tikzpicture}[remember picture,overlay]%
     \node[anchor=south, xshift=0pt, yshift=20pt] at (current page.south)%
     {\fbox{\parbox{\dimexpr\textwidth-\fboxsep-\fboxrule\relax}{\copyrighttext}}};%
     \end{tikzpicture}%
}

\title{The ADUULM-360 Dataset - \\ A Multi-Modal Dataset for Depth Estimation in Adverse Weather}
\author{Markus Schön, Jona Ruof, Thomas Wodtko, Michael Buchholz, and Klaus Dietmayer%
\thanks{This work was financially supported by the Federal Ministry of Education
and Research (BMBF) (project AUTOtech.agil, FKZ 01IS22088W) and the State
Ministry of Economic Affairs Baden-Württemberg (project U-Shift II, AZ 3-
433.62-DLR/60).}%
\thanks{All authors are with
the Institute of Measurement, Control, and Microtechnology
of the University of Ulm, 89081, Ulm, Germany {\tt\footnotesize \{firstname\}.\{lastname\}@uni-ulm.de}}%
}

\begin{document}

\maketitle

\begin{abstract}
Depth estimation is an essential task toward full scene understanding since it allows the projection of rich semantic information captured by cameras into 3D space.
While the field has gained much attention recently, datasets for depth estimation lack scene diversity or sensor modalities.
This work presents the ADUULM-360 dataset, a novel multi-modal dataset for depth estimation.
The ADUULM-360 dataset covers all established autonomous driving sensor modalities, cameras, lidars, and radars.
It covers a frontal-facing stereo setup, six surround cameras covering the full 360-degree, two high-resolution long-range lidar sensors, and five long-range radar sensors.
It is also the first depth estimation dataset that contains diverse scenes in good and adverse weather conditions.
We conduct extensive experiments using state-of-the-art self-supervised depth estimation methods under different training tasks, such as monocular training, stereo training, and full surround training.
Discussing these results, we demonstrate common limitations of state-of-the-art methods, especially in adverse weather conditions, which hopefully will inspire future research in this area.
Our dataset, development kit, and trained baselines are available at \url{https://github.com/uulm-mrm/aduulm_360_dataset}.
\end{abstract}
\section{Introduction}
\copyrightnotice
Autonomous driving is a challenging task that requires sophisticated methods in many fields, such as perception, planning, and control.
Perception systems must deliver an accurate and robust model of the vehicle's environment, including detection and localization of all traffic participants and essential regions,~\eg, drivable space and sidewalks.
To achieve this, modern perception systems rely on high-quality sensors of multiple modalities, such as cameras, lidar, and radar sensors.
Using multiple modalities becomes especially important in adverse weather conditions where a single modality may be disturbed or fail~\cite{liu2020computing}.

Deep learning has made a lasting change to the field of perception, with trained deep neural networks achieving outstanding performance on almost all perception tasks.
One key takeaway of the last decade of deep learning research is that model performance highly depends on the amount and quality of training data.
In general, a large amount of high-quality and diverse data is beneficial for the performance and generalization ability of trained models~\cite{simon2024more}.
Acquiring high-quality data is not easy, and the manual annotation process can be cumbersome and error-prone.
Research in some fields shifted towards self-supervised methods to soften the dataset generation process, which does not require hand-labeled annotations but uses the acquired data directly for network training.
One such task is camera-based depth estimation, where manual annotation of per-pixel accurate ground truth is nearly impossible.
\begin{figure}
    \centering
    \includegraphics[width=\columnwidth]{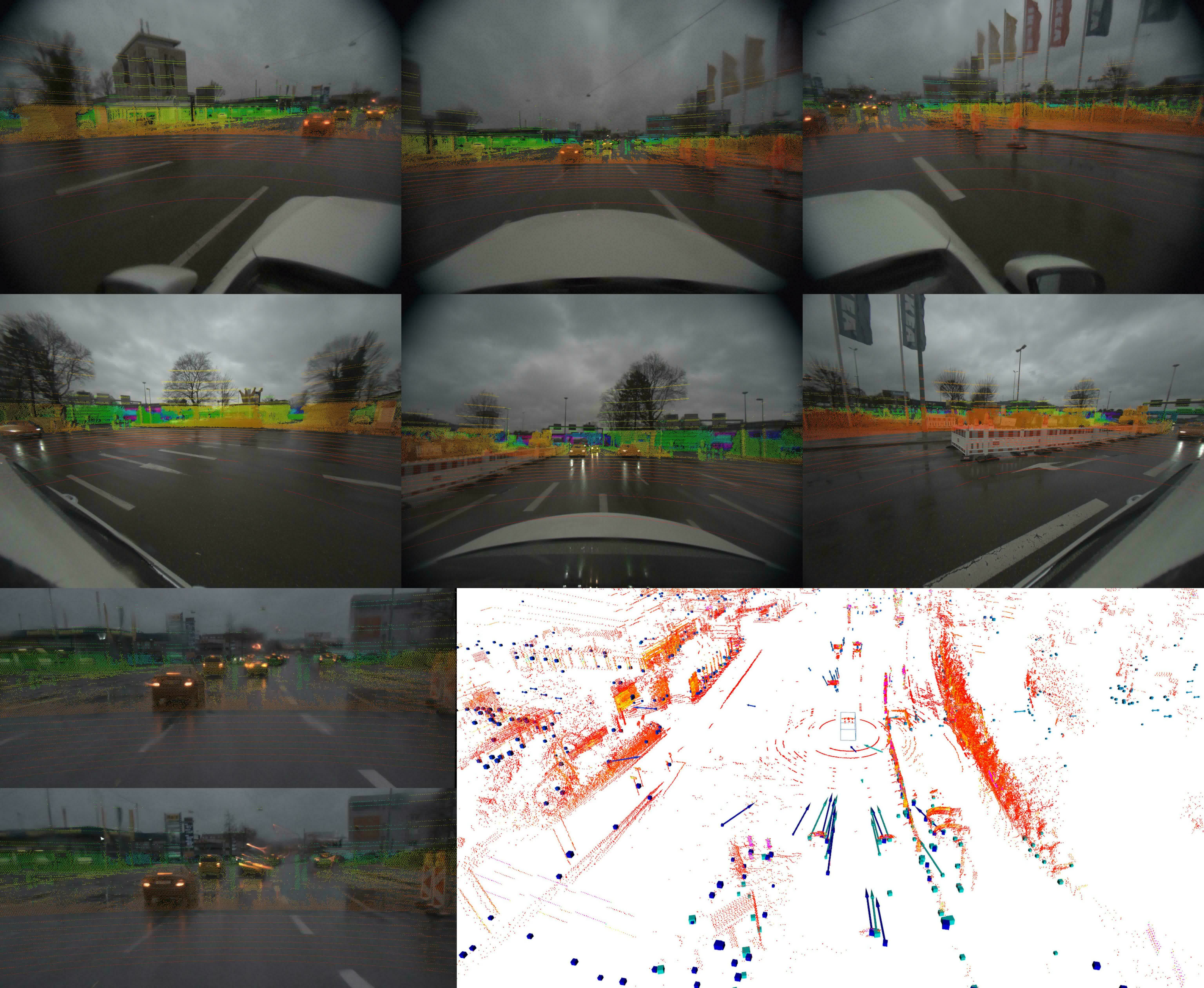}
    \caption{An example from the ADUULM-360 dataset in light rain conditions. The top shows lidar projections in the six surround camera views, the bottom left in the stereo views, and the bottom right shows the lidar and radar point clouds.}
    \label{fig:data_example}
\end{figure}

Self-supervised depth estimation methods~\cite{zhou2017unsupervised, godard2019digging, guizilini20203d, zhao2022monovit, zhang2023lite} reformulate the task as an image-to-image synthesis problem and train the model using a photometric loss.
This only requires stereo images or monocular video sequences during training, which enables the use of large amounts of unlabeled data.
Trained models are validated using lidar measurements projected to the image space, which only requires accurate sensor calibration.
Research in the field made significant progress in recent years with modern transformer-based architectures~\cite{zhao2022monovit, zhang2023lite, wei2023surrounddepth} achieving high performance on standard benchmarks such as KITTI~\cite{geiger2013vision} and DDAD~\cite{guizilini20203d}.

\begin{figure*}
    \vspace{0.15cm}
    \centering
    \includegraphics[width=\textwidth, height=5cm]{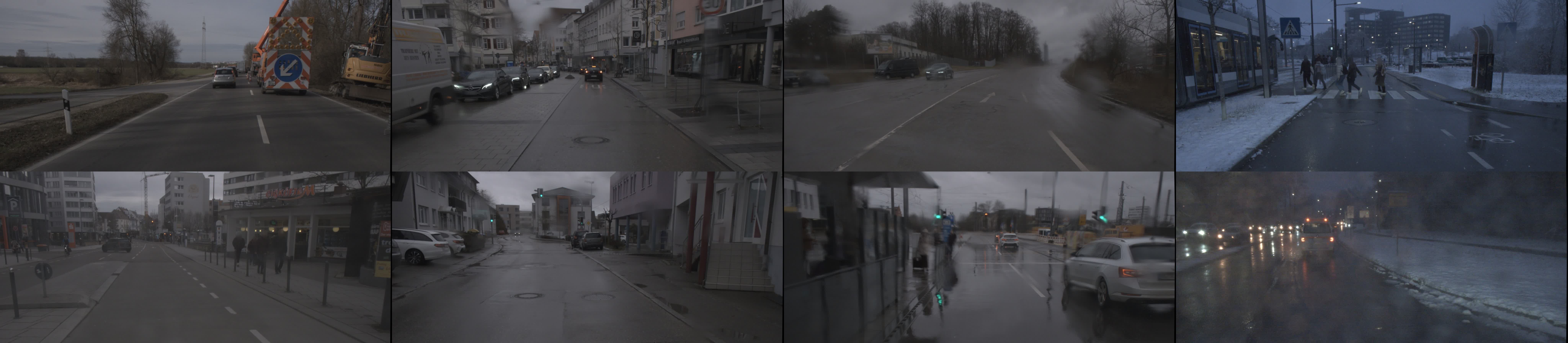}
    \caption{Dataset samples of the frontal facing left stereo camera at different lighting and weather conditions. The first columns show examples of good weather; the middle two columns show examples of lighter and heavier rain. The last column shows examples of snowing conditions.}
    \label{fig:data_adverse}
    \vspace{-0.15cm}
\end{figure*}

While both KITTI and DDAD provide data in good weather conditions, up to this point, depth estimation research for adverse weather conditions mainly relies on simulating adverse weather data by augmenting good weather data~\cite{saunders2023self, kong2023robodepth}.
While this approach is reasonable due to the lack of real-world adverse weather data availability, it is unclear if approaches trained on simulated adverse data achieve similar performance in real-world conditions.
Autonomous driving datasets that do contain adverse weather scenes, such as our previously published ADUULM dataset~\cite{pfeuffer2020aduulm}, nuScenes~\cite{caesar2020nuscenes}, or ZOD~\cite{alibeigi2023zenseact}, could, in theory, be used for self-supervised depth estimation.
However, they lack specific modalities, such as stereo cameras and baseline results for depth estimation models.
Thus, these datasets are largely unused by the depth estimation community.

To fill this gap, we introduce the ADUULM-360 dataset, a multi-modal dataset for depth estimation in good and adverse weather conditions.
The ADUULM-360 dataset provides high-quality and diverse data from multiple cameras, lidar sensors, and radar sensors, covering the full 360 degrees of the vehicle's surroundings.
Fig.~\ref{fig:data_example} shows one dataset sample in light rain conditions.
The sensor suite consists of six surround cameras (top), two frontal-facing stereo cameras (bottom left), and point clouds of two long-range lidar sensors and five long-range radar sensors (bottom right).
Besides light rain, the dataset also covers more extreme adverse weather conditions such as heavy rain and snow.
Furthermore, it covers different times of day and lighting conditions at a large scale.
In total, the dataset provides 3.5 hours of captured sensor data, consisting of approximately 1M camera images, 250k lidar point clouds, and 800k radar point clouds, of which approximately 20\% was captured during adverse weather.
Fig.~\ref{fig:data_adverse} shows the data diversity based on samples of the left stereo camera view.

In addition to the dataset itself, we define three benchmarks: 
Self-supervised monocular depth estimation, stereo depth estimation, and full-surround depth estimation.
We provide baselines on each benchmark for different self-supervised depth estimation methods,~\ie, MonoDepth2~\cite{godard2019digging}, MonoViT~\cite{zhao2022monovit}, Lite-Mono~\cite{zhang2023lite}, and SurroundDepth~\cite{wei2023surrounddepth}, and discuss limitations of current approaches.
Furthermore, we investigate the effect of adverse weather on these methods and show that simply training with both good and adverse weather samples does not necessarily improve adverse weather performance.

In summary, our main contributions are the following:
\begin{itemize}
    \item We release the ADUULM-360 dataset, a multi-modal dataset for depth estimation in good and adverse weather conditions. The dataset comprises long sequences covering a rich set of driving scenarios and weather conditions.
    \item The data is collected using measurements of 15 sensors, eight cameras, two lidar sensors, and five radar sensors, covering the entire 360-degree vehicle surroundings and an extensive range of up to 250 m.
    \item We benchmark state-of-the-art self-supervised depth estimation methods on our dataset to provide a set of baselines to encourage future research to use the dataset in their experiments.
    \item We investigate the performance of state-of-the-art methods on our dataset and discuss common limitations of these methods under different weather conditions.
\end{itemize}
\section{Related Work}
\subsection{Depth Estimation Datasets}
The KITTI~\cite{geiger2013vision} dataset is the de-facto standard benchmark when evaluating depth estimation model performance for autonomous driving applications.
While it provides many video sequences and stereo images of frontal-facing cameras, it lacks surrounding views.
It also provides sparse lidar ground truth for a relatively low range.
The DDAD~\cite{guizilini20203d} dataset was introduced to address these limitations by providing surrounding camera views and ground truth from a more modern lidar sensor with a maximum range of up to 250~\si{m}.
Compared to KITTI, DDAD provides a more realistic and challenging dataset.
However, DDAD does not provide stereo cameras, which several methods use as an alternative to monocular training based on video sequences.
Furthermore, DDAD lacks adverse weather data, making trained approaches unsafe to use under real-world adverse weather conditions.
In contrast, our dataset provides diverse data from good and adverse weather scenes and includes various sensor modalities, including stereo and surround cameras.
The nuScenes~\cite{caesar2020nuscenes} dataset is used occasionally for self-supervised depth estimation~\cite{guizilini20203d, wei2023surrounddepth}.
While nuScenes provides diverse data in good and adverse weather conditions, the dataset lacks stereo cameras, and the lidar used as ground truth only provides 32 layers in a close range of up to 80~\si{m}.
In contrast, our dataset provides higher-resolution lidar data with a maximum range of up to 250~\si{m}.

\subsection{Self-supervised Depth Estimation Methods}
Garg~\etal~\cite{garg2016unsupervised} were the first to train a depth estimation network in a self-supervised way using stereo images during training.
They reformulated the problem as an image-to-image synthesis problem by warping the right camera image of a stereo camera pair into the left image using the predicted depth and minimizing the photometric error between the left and the warped right camera image.
Zhou~\etal~\cite{zhou2017unsupervised} were the first to use monocular video sequences instead of stereo images.
The idea is similar to stereo images, but a separate pose network learns the relative transformation between the video sequence images, which is then used to synthesize the warped images.
Godard~\etal~\cite{godard2019digging} boosted performance significantly by introducing an auto-masking strategy and using the minimum photometric error to tackle occlusions.
Many works~\cite{guizilini20203d, zhao2022monovit, zhang2023lite} are based on their self-supervised framework.
PackNet~\cite{guizilini20203d} uses symmetric packing and unpacking blocks to improve the representation ability of the model to encode fine-grained details.
MonoViT~\cite{zhao2022monovit} and Lite-Mono~\cite{zhang2023lite} both use transformer-based architectures to achieve state-of-the-art performance in the heavy and lightweight architecture domains.
We provide baselines for MonoDepth2, MonoViT, and Lite-Mono trained on our ADUULM-360 dataset.
Furthermore, we investigate the effects of adverse weather on the different methods.

\subsection{Full-Surround Depth Estimation Methods}
Guizilini~\etal~\cite{guizilini2022full} were the first to extend the self-supervised depth formulation to a 360-degree camera setup.
Their approach uses spatio-temporal contexts and pose consistency constraints to learn a single network that produces a full 360-degree point cloud based on surround view images.
Wei~\etal~\cite{wei2023surrounddepth} propose a cross-view transformer to effectively fuse the information from multiple views in the network, which improves depth consistency across camera views.
We train their method, SurroundDepth, on the ADUULM-360 dataset to provide a strong baseline for full-surround depth estimation.
\section{Dataset}
This section details the data collection and dataset generation process of the ADUULM-360 dataset.
First, the data collection process is described.
Then, the sensor setup and sensor calibration are explained.
Afterward, the data processing pipeline is detailed, which prepares the raw sensor data for self-supervised depth estimation training.
Lastly, the anonymization method is described, which ensures data privacy.

\subsection{Data Collection}
The ADUULM-360 dataset was collected using sensor data from a Mercedes Benz S-Class test vehicle, which has a permit to drive on public roads in Germany.
The data was collected in and around the city of Ulm, Germany.
Driving routes were carefully chosen to ensure data diversity in terms of location (city center, residential area, rural area, highway), time of day (morning, afternoon, evening), and weather condition (sunny, cloudy, rainy, and snowy).
A total of 60 sequences of various lengths between one and ten minutes were manually chosen based on the drives, with a total driving time of 3.5 hours.
In total, the dataset comprises approximately 1M camera images, 250k lidar point clouds, and 800k radar point clouds, of which approximately 20\% was captured in adverse weather conditions.

\subsection{Sensor Setup}
\begin{figure}
\vspace{0.15cm}
\input{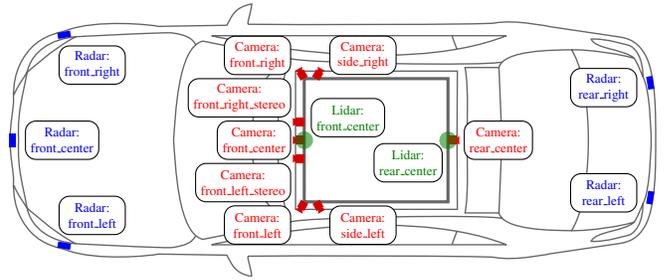}
\vspace{-0.6cm}
\caption{Overview of the sensor setup used for the ADUULM-360 dataset. Camera sensors are depicted in red, lidar sensors in green, and radar sensors in blue.}
\label{fig:sensor_setup}
\end{figure}

Fig.~\ref{fig:sensor_setup} shows the sensor setup of the test vehicle.
To realize a visual 360-degree coverage of the environment, the test vehicle is equipped with eight identical, high-resolution cameras mounted on the roof.
Most notably, all cameras have overlapping \glspl{fov} with at least two different cameras, which facilitates using,~\eg, surround depth estimation methods.
Two rear-facing cameras were mounted on either side of the vehicle to primarily cover its close proximity.
These are equipped with fisheye lenses with a $\approx 190^{\circ}$ \gls{fov}, which covers the entire vehicle side up to the rear-view mirror. 
Additionally, the side cameras are supported by another back-facing camera at the rear-center of the vehicle, equipped with wide-angle optics with a $\approx 100^{\circ}$ \gls{fov}.
On the front of the vehicle, three cameras with the same wide-angle optics cover the environment to the left, right, and center.
Finally, in the front center, a stereo camera pair with $\approx 60^{\circ}$ \gls{fov} lenses provides additional coverage at greater distances.
All images are captured with $2\times$ sensor binning at a resolution of $2048\times1500\,\textrm{px}$.
All cameras perform automatic exposure time adjustment with a maximum exposure time of $\SI{3}{ms}$, sensor gain adjustment, and a gamma correction with $\gamma = 0.5$.
Since many processing steps, such as depth estimation, require synchronized images, all cameras are triggered in unison at $\SI{10}{Hz}$ using an onboard, high-accuracy \gls{ptp} grandmaster clock unit.

In addition to the cameras, which constitute the main focus of the sensor setup, the test vehicle is equipped with lidar and radar sensors and an \gls{imu} with \gls{rtk}. 
The two lidar sensors are mounted on the roof, above the cameras, at the front and rear center of the vehicle, respectively. 
This results in a large overlapping \gls{fov} between the two lidars and between the lidars and the cameras. 
Each lidar provides a point cloud with 64 layers with an angular resolution of $0.2^\circ$ or 1800 points per layer. 
Both sensors are triggered at 10 Hz and synchronized via \gls{ptp} to the onboard clock unit and, thus, to the cameras.
Five off-the-shelve automotive \gls{fmcw} radar sensors provide a third sensor modality capturing data at $\SI{12.5}{Hz}$. 
Three radars are mounted at the front of the vehicle, one in the center, the left, and the right corner, respectively.
Two radars are mounted back-facing on the vehicle's rear to cover the neighboring lanes.
Note that the data from the radar sensors is not used in our benchmark results. 
However, it is still included in the dataset and may be helpful for future work.

\subsection{Sensor Calibration}
All sensors are calibrated intrinsically and extrinsically to ensure data alignment.
For monocular cameras, checkerboards are used for intrinsic calibration to estimate distortion and the pinhole camera matrix parameters~\cite{zhang2000flexible}.
A Scaramuzza model~\cite{scaramuzza2006toolbox} is estimated based on checkerboard recordings for fisheye cameras.
Since most depth estimation methods rely on pinhole camera models, the fisheye images are projected to a virtual pinhole model using the Scaramuzza model and a set output focal length and optical center.
For stereo cameras, the images are rectified to allow stereo depth estimation and cropped to a size of $2048\times896\,\textrm{px}$ to only include valid pixels after rectification.
Cameras, lidars, and radars are calibrated extrinsically using the target-based approaches described in~\cite{horn2023extrinsic} and~\cite{kummerle2020multimodal}.
We refer the reader to these publications for details on extrinsic calibration.

\subsection{Data Processing}
All camera images are undistorted, stereo camera images are rectified, and fisheye images are projected into pinhole images.
The raw lidar measurements cannot be used as ground truth directly since they are prone to ego-motion distortion.
The ego vehicle localization measurements provided by the \gls{imu} are filtered using an unscented Kalman filter.
The filtered ego-motion data is used to compensate the lidar point clouds for the ego-motion of the test vehicle.
Both lidar sensors' ego-motion compensated point clouds are combined and transformed into the camera images using extrinsic calibration.
Following common practice~\cite{geiger2013vision}, we use the closest lidar point if two or more lidar points fall into the same image pixel.
Lidar point clouds suffer from clutter and other effects in adverse weather conditions, such as reflections of wet road surfaces.
Since these clutter points do not provide accurate measurements of the environment, they cannot be used as ground truth since they would corrupt metric calculations.
Hence, these clutter points must be filtered out for ground truth generation.
Clutter detection in adverse weather is still an active research topic.
We settled on two simple but effective approaches.
First, we filter out points with a low-intensity measurement below one.
This reduces the number of noisy measurements mainly present in snowy conditions.
Second, we use PathWork++~\cite{lee2022patchworkpp} to estimate a ground plane in the compensated lidar point cloud and filter out all points below the ground plane to handle ground reflections in the lidar point cloud.

\subsection{Data Privacy}
Data privacy is essential to datasets that contain sensible data, such as the faces of persons and vehicle license plates.
To protect personal data, we use an anonymization tool~\cite{dashcam_anonymizer} based on the image object detection model YOLOv8~\cite{Jocher_Ultralytics_YOLO_2023} to anonymize all camera images.
The detector is trained to recognize faces and license plates.
We apply the detector with a low detection confidence threshold of $0.1$ to reduce false negatives.
All detected regions are blurred using a Gaussian blur with a radius of 31.
\begin{table*}[t]
\centering
  \vspace{0.2cm}
  \caption{Performance comparison of self-supervised monocular depth estimation methods on the different ADUULM-360 test splits. \\ \textbf{M}: trained with monocular videos, \textbf{S}: trained with synchronized stereo pairs, \textbf{Best}: \textbf{boldface}, \textbf{Second best}: \underline{underlined}.}
  \resizebox{\textwidth}{!}{ %
  \begin{tabular}{c|c|c|c|c|c|c|c|c|c}
    \toprule
    Split & Method &Train &\cellcolor{red!25}AbsRel$\downarrow$ &\cellcolor{red!25}SqRel$\downarrow$ &\cellcolor{red!25}RMSE$\downarrow$ &\cellcolor{red!25}RMSElog$\downarrow$ &\cellcolor{blue!25}$\delta<1.25\uparrow$ &\cellcolor{blue!25}$\delta<1.25^2\uparrow$ &\cellcolor{blue!25}$\delta<1.25^3\uparrow$   \\
    \midrule
    \multirow{8}{*}{Good} & MonoDepth2~\cite{godard2019digging} & M & 0.300 & \underline{20.638} & 17.077 & 0.362 & \underline{0.749} & 0.884 & 0.935 \\
    & MonoViT~\cite{zhao2022monovit} & M & \underline{0.290} & 22.103 & \underline{16.656} & \textbf{0.345} & \textbf{0.780} & \textbf{0.893} & \textbf{0.939} \\
    & Lite-Mono-Small~\cite{zhang2023lite} & M & \textbf{0.274} & \textbf{15.689} & \textbf{16.596} & \underline{0.352} & 0.736 & \underline{0.889} & \underline{0.938} \\
    & Lite-Mono~\cite{zhang2023lite} & M & 0.311 & 23.121 & 17.511 & 0.363 & 0.747 & 0.885 & 0.936 \\
    \cmidrule{2-10}
    & MonoDepth2~\cite{godard2019digging} & S & \underline{0.166} & 4.637 & \underline{11.390} & \underline{0.294} & \underline{0.817} & \underline{0.913} & \underline{0.951} \\
    & MonoViT~\cite{zhao2022monovit} & S & \textbf{0.152} & \textbf{2.902} & \textbf{10.758} & \textbf{0.283} & \textbf{0.820} & \textbf{0.917} & \textbf{0.955} \\
    & Lite-Mono-Small~\cite{zhang2023lite} & S & 0.178 & \underline{3.079} & 11.621 & 0.316 & 0.785 & 0.903 & 0.944 \\
    & Lite-Mono~\cite{zhang2023lite} & S & 0.180 & 4.583 & 12.171 & 0.321 & 0.800 & 0.908 & 0.945 \\
    \midrule
    \midrule
    \multirow{8}{*}{Light} & MonoDepth2~\cite{godard2019digging} & M & 0.255 & 8.360 & 14.291 & 0.340 & \underline{0.708} & \underline{0.876} & 0.935 \\
    & MonoViT~\cite{zhao2022monovit} & M & \textbf{0.230} & \textbf{7.101} & \textbf{13.293} & \textbf{0.318} & \textbf{0.735} & \textbf{0.891} & \textbf{0.944} \\
    & Lite-Mono-Small~\cite{zhang2023lite} & M & 0.268 & 9.120 & 14.903 & 0.358 & 0.667 & 0.859 & 0.928 \\
    & Lite-Mono~\cite{zhang2023lite} & M & \underline{0.248} & \underline{7.189} & \underline{14.118} & \underline{0.339} & 0.699 & 0.873 & \underline{0.936} \\
    \cmidrule{2-10}
    & MonoDepth2~\cite{godard2019digging} & S & 0.226 & 7.717 & 14.387 & \underline{0.342} & \underline{0.742} & \underline{0.879} & \underline{0.932} \\
    & MonoViT~\cite{zhao2022monovit} & S & \textbf{0.208} & \textbf{4.227} & \textbf{12.253} & \textbf{0.313} & \textbf{0.749} & \textbf{0.890} & \textbf{0.942} \\
    & Lite-Mono-Small~\cite{zhang2023lite} & S & \underline{0.217} & \underline{4.425} & \underline{13.510} & 0.355 & 0.705 & 0.858 & 0.920 \\
    & Lite-Mono~\cite{zhang2023lite} & S & 0.227 & 7.093 & 14.283 & 0.353 & 0.724 & 0.869 & 0.927 \\
    \midrule
    \midrule
    \multirow{8}{*}{Heavy} & MonoDepth2~\cite{godard2019digging} & M & 0.506 & 37.519 & 22.234 & 0.488 & 0.571 & 0.774 & 0.866 \\
    & MonoViT~\cite{zhao2022monovit} & M & \textbf{0.395} & 27.152 & \underline{20.102} & \textbf{0.427} & \textbf{0.649} & \textbf{0.831} & \textbf{0.903} \\
    & Lite-Mono-Small~\cite{zhang2023lite} & M & \underline{0.423} & \textbf{20.800} & \textbf{20.032} & 0.470 & 0.545 & 0.774 & 0.876 \\
    & Lite-Mono~\cite{zhang2023lite} & M & 0.438 & \underline{25.118} & 20.698 & \underline{0.462} & \underline{0.574} & \underline{0.788} & \underline{0.881} \\
    \cmidrule{2-10}
    & MonoDepth2~\cite{godard2019digging} & S & 0.274 & 12.912 & 18.433 & \underline{0.422} & \underline{0.662} & \underline{0.824} & \underline{0.896} \\
    & MonoViT~\cite{zhao2022monovit} & S & \textbf{0.220} & \textbf{4.525} & \textbf{14.562} & \textbf{0.368} & \textbf{0.681} & \textbf{0.843} & \textbf{0.911} \\
    & Lite-Mono-Small~\cite{zhang2023lite} & S & \underline{0.257} & \underline{6.097} & \underline{16.962} & 0.452 & 0.605 & 0.789 & 0.872 \\
    & Lite-Mono~\cite{zhang2023lite} & S & 0.283 & 12.209 & 18.300 & 0.441 & 0.641 & 0.815 & 0.889 \\
    \bottomrule
  \end{tabular}
  }
  \label{tab:monocular}
\end{table*}

\section{Tasks and Metrics}
The ADUULM-360 dataset provides multi-modal data that may be used for many tasks.
We focus this work on self-supervised depth estimation and define three tasks to which we provide baseline models in the experiments section:
\begin{enumerate}
    \item Self-supervised Monocular Depth Estimation: Estimation of per-pixel depth using video sequence images from the left stereo camera only.
    \item Self-supervised Stereo Depth Estimation: Estimation of per-pixel depth using camera images from the left and right stereo camera.
    \item Self-supervised Full-surround Depth Estimation: Estimation of per-pixel depth for all six monocular cameras around the vehicle to cover the complete 360-degree environment.
\end{enumerate}
We use the same dataset splits across all tasks.
The dataset contains $\approx\text{130k}$ samples, each providing synchronized camera images and projected lidar point clouds used as ground truth.
The dataset is split into two training sets and three test sets.
The main training set contains $\approx\text{90k}$ samples captured in good weather conditions.
The second training set contains $\approx\text{13k}$ samples captured in adverse weather conditions that can be used to improve the generalization ability of models for adverse weather conditions.
Each test set contains 9k samples and is separated based on weather conditions.
The first test set contains only good weather samples, the second one contains samples with lighter adverse weather conditions, such as light rain, and the third one contains samples with heavier adverse weather conditions, such as heavy rain or snowfall.

Given a depth prediction $y$ and depth ground truth $y^*$, we report the following depth metrics~\cite{eigen2014depth} for all tasks:
\vspace{0.5em}
\begin{itemize}
    \setlength\itemsep{0.5em}
    \item AbsRel:  $\frac{1}{N}\sum_{i=1}^N|y_i - y_i^*| / y_i^*$
    \item SqRel:  $\frac{1}{N}\sum_{i=1}^N||y_i - y_i^*||^2 / y_i^*$
    \item RMSE:  $\sqrt{\frac{1}{N}\sum_{i=1}^N||y_i - y_i^*||^2}$
    \item RMSElog:  $\sqrt{\frac{1}{N}\sum_{i=1}^N||\log y_i - \log y_i^*||^2}$
    \item Threshold Metrics $\delta < T_n$: $\frac{1}{N}\left|\left\{\max\left(\frac{y_i}{y_i^*},\frac{y_i^*}{y_i}\right) < T_n\right\}\right|$
\end{itemize}
\vspace{0.5em}
where $N$ is the number of valid pixels in $y^*$ and $T_n$ are a number of accuracy thresholds.
For full-surround depth estimation, following~\cite{guizilini2022full}, metrics are calculated for each camera view separately and averaged to give a final score.
All metrics are calculated for depth values up to $\SI{200.}{m}$.
\section{Experimental Results}
This section provides experimental results on the ADUULM-360 dataset.
First, baselines of state-of-the-art methods for each task are evaluated.
Then, an ablation study is carried out using adverse weather data during training.
Finally, the results are discussed, and common limitations of the provided baselines that might be tackled in future research using the ADUULM-360 dataset are pointed out.

\subsection{Baseline Results}
\label{sec:baselines}
For the tasks of self-supervised monocular depth estimation and self-supervised stereo depth estimation, we train the state-of-the-art methods MonoDepth2~\cite{godard2019digging}, MonoViT~\cite{zhao2022monovit}, and Lite-Mono~\cite{zhang2023lite} on the ADUULM-360 dataset.
All models are trained using the good weather training set in two settings: video sequence training and rectified stereo image training.
Training settings are chosen according to the training settings of each method on the KITTI~\cite{geiger2013vision} dataset.
We refer the reader to the respective paper for specific implementation details.
We set the model input image size to $640\times280\,\textrm{px}$ to match the aspect ratio of the stereo images of the ADUULM-360 dataset.
Furthermore, we filter out static frames based on the ego vehicle velocity for video sequence training since static frames corrupt photometric loss calculation.
In particular, we use the velocity provided by the ego-motion data and set a velocity threshold of $\SI{1}{m/s}$ to decide if a frame is considered static.
Since video-sequence training suffers from scale ambiguity, we use median depth rescaling for these baselines~\cite{godard2019digging}.
Table~\ref{tab:monocular} shows the quantitative results of the models on the three test sets of the ADUULM-360 dataset.
For monocular video sequence training, Lite-Mono-Small provides the best results in good weather conditions, while MonoViT offers the best results on both adverse weather test splits.
For stereo training, MonoViT provides the best results across all test splits, with Lite-Mono being second best, providing slightly better results than MonoDepth2.
As expected, the performance across all methods decreases under adverse weather conditions compared to good weather conditions.
One exception is the video sequence training, where results on the light adverse weather test set are better compared to the good weather test set.
This will be discussed in Section~\ref{sec:discussion}.

For self-supervised full-surround depth estimation, we train the state-of-the-art method SurroundDepth~\cite{wei2023surrounddepth} using the same training settings used by the authors when training on the DDAD~\cite{guizilini20203d} dataset.
We adjust the model input image size to $640\times480\,\textrm{px}$ to match the aspect ratio of the surround view images of the ADUULM-360 dataset.
Since the surround view images capture parts of the ego-vehicle, we manually annotate occlusion masks to filter these regions out for photometric loss calculation.
Quantitive results for this method on the three test splits of the ADUULM-360 dataset can be found in Table~\ref{tab:surround}.
Similar to monocular video sequence training, the average error under light adverse weather conditions is slightly lower than under good weather conditions.
For heavier adverse weather conditions, as expected, the average error increases.
However, the increase is more pronounced than in the monocular case and depends on the camera view, which will be discussed in Section~\ref{sec:discussion}.

\begin{table}[t!]
    \centering
    \vspace{0.2cm}
    \caption{Performance of the self-supervised full-surround depth estimation method SurroundDepth~\cite{wei2023surrounddepth} on the different ADUULM-360 test splits. \textbf{FC}: Front Center, \textbf{FL}: Front Left, \textbf{FR}: Front Right, \textbf{SL}: Side Left, \textbf{SR}: Side Right, \textbf{RC}: Rear Center}
    {
    \small
    \setlength{\tabcolsep}{0.3em}
    \begin{tabular}{l|cccccc|c}
    \toprule
    Split & \multicolumn{6}{c}{\cellcolor{red!25}AbsRel$\downarrow$} & \multirow{2}{*}{} \\
    \cmidrule{2-8}
     & FC & FL & FR & SL & SR & RC & Avg \\
    \midrule
    Good & 0.561 & 0.425 & 0.579 & 0.463 & 0.345 & 0.515 & 0.481 \\
    Light & 0.539 & 0.371 & 0.591 & 0.369 & 0.374 & 0.411 & 0.443 \\
    Heavy & 0.677 & 0.453 & 0.665 & 0.438 & 0.428 & 0.537 & 0.533 \\
    \bottomrule
    \end{tabular}
    }
    \label{tab:surround}
\end{table}

\subsection{Ablation Study}
\label{sec:ablation}
We perform an ablation study to investigate whether using adverse weather data during training improves the generalization under these conditions.
Using the adverse training set, we perform two training strategies.
First, the adverse training set is used in conjunction with the good weather training set.
Second, the adverse training set is used to fine-tune a model that was previously trained on the good weather training set.
For both strategies, we use MonoViT and train both strategies using stereo images as described in Section~\ref{sec:baselines}.
The trained baseline is used as initialization for the fine-tuning strategy.
Results are shown in Table~\ref{tab:adverse}.
As can be seen, training using both datasets in conjunction leads to slightly better performance in good and heavy weather conditions, while performance in light weather conditions drops slightly.
When performing fine-tuning, adverse weather performance worsens compared to the baseline.
This is further discussed in the next section.

\begin{table}[t]
\centering
  \caption{Comparison of different strategies for using adverse weather data during training. \textbf{Train} uses adverse weather data during training, while \textbf{Fine} uses adverse weather data for fine-tuning only. All experiments use MonoViT~\cite{zhao2022monovit} trained with synchronized stereo image pairs.}
  \begin{tabular}{c|c|c|c|c|c}
    \toprule
    Split & Train & Fine &\cellcolor{red!25}AbsRel$\downarrow$ &\cellcolor{red!25}RMSE$\downarrow$ &\cellcolor{blue!25}$\delta<1.25\uparrow$ \\
    \midrule
    \multirow{3}{*}{Good} & & & \textbf{0.152} & \underline{10.758} & \underline{0.820} \\
    & \cmark & & \textbf{0.152} & \textbf{10.475} & \textbf{0.833} \\
    & & \cmark & \underline{0.168} & 10.815 & 0.804 \\
    \midrule
    \midrule
    \multirow{3}{*}{Light} & & & \textbf{0.208} & \textbf{12.253} & \textbf{0.749} \\
    & \cmark & & \underline{0.235} & \underline{13.354} & \underline{0.746} \\
    & & \cmark & 0.293 & 15.219 & 0.704 \\
    \midrule
    \midrule
    \multirow{3}{*}{Heavy} & & & \textbf{0.220} & \underline{14.562} & \underline{0.681} \\
    & \cmark & & \underline{0.229} & \textbf{14.043} & \textbf{0.699} \\
    & & \cmark & 0.284 & 15.058 & 0.660 \\
    \bottomrule
  \end{tabular}
  \label{tab:adverse}
\end{table}

\subsection{Discussion}
\label{sec:discussion}
\begin{figure*}[t!]
    \centering
    \subfloat[Example from the good weather test split.]{%
      \noindent\includegraphics[width=0.3\textwidth]{"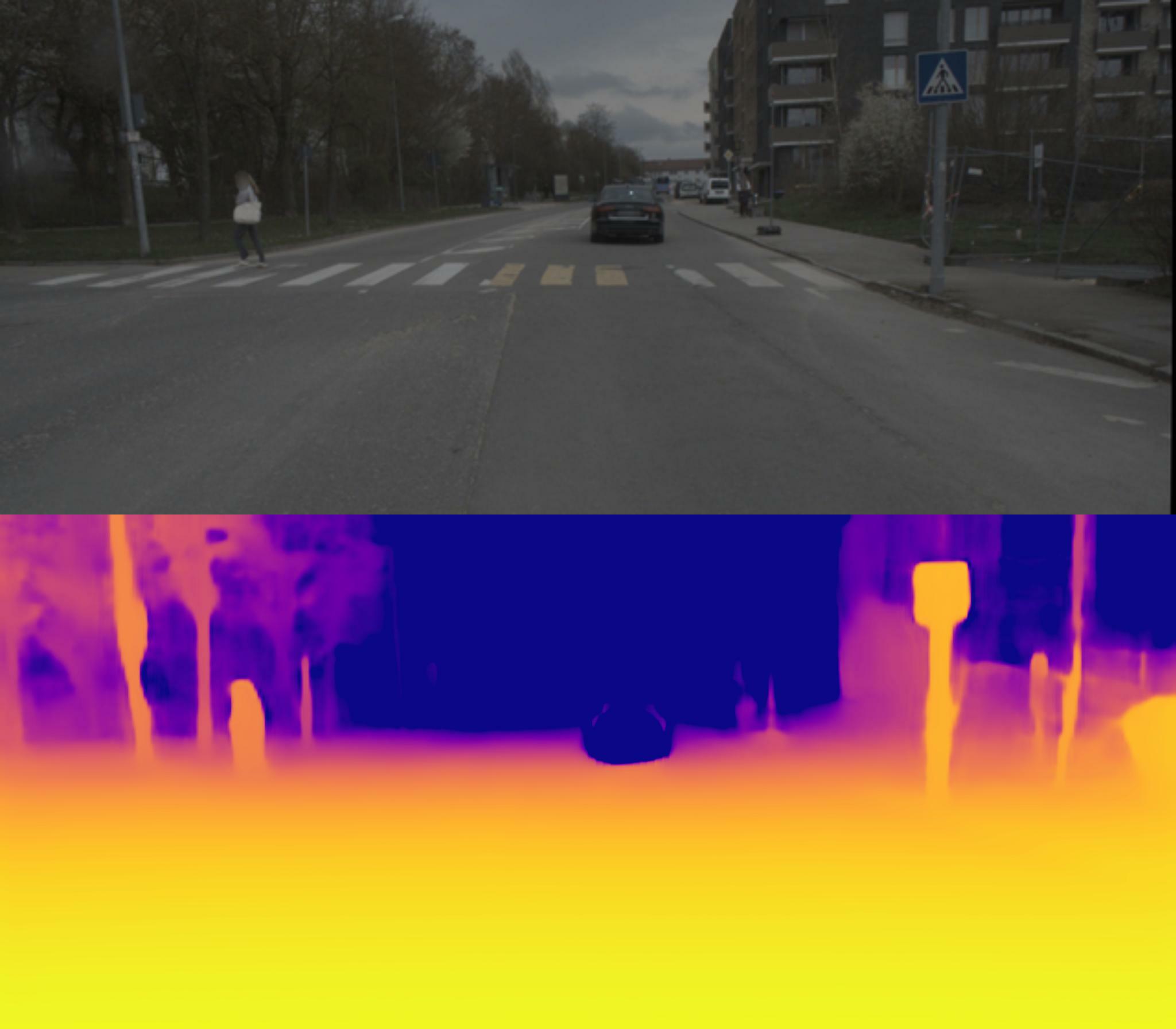"}%
      \label{fig:qualitative_comp_good}
    }
    \hfill
    \subfloat[Example from the light weather test split.]{%
      \noindent\includegraphics[width=0.3\textwidth]{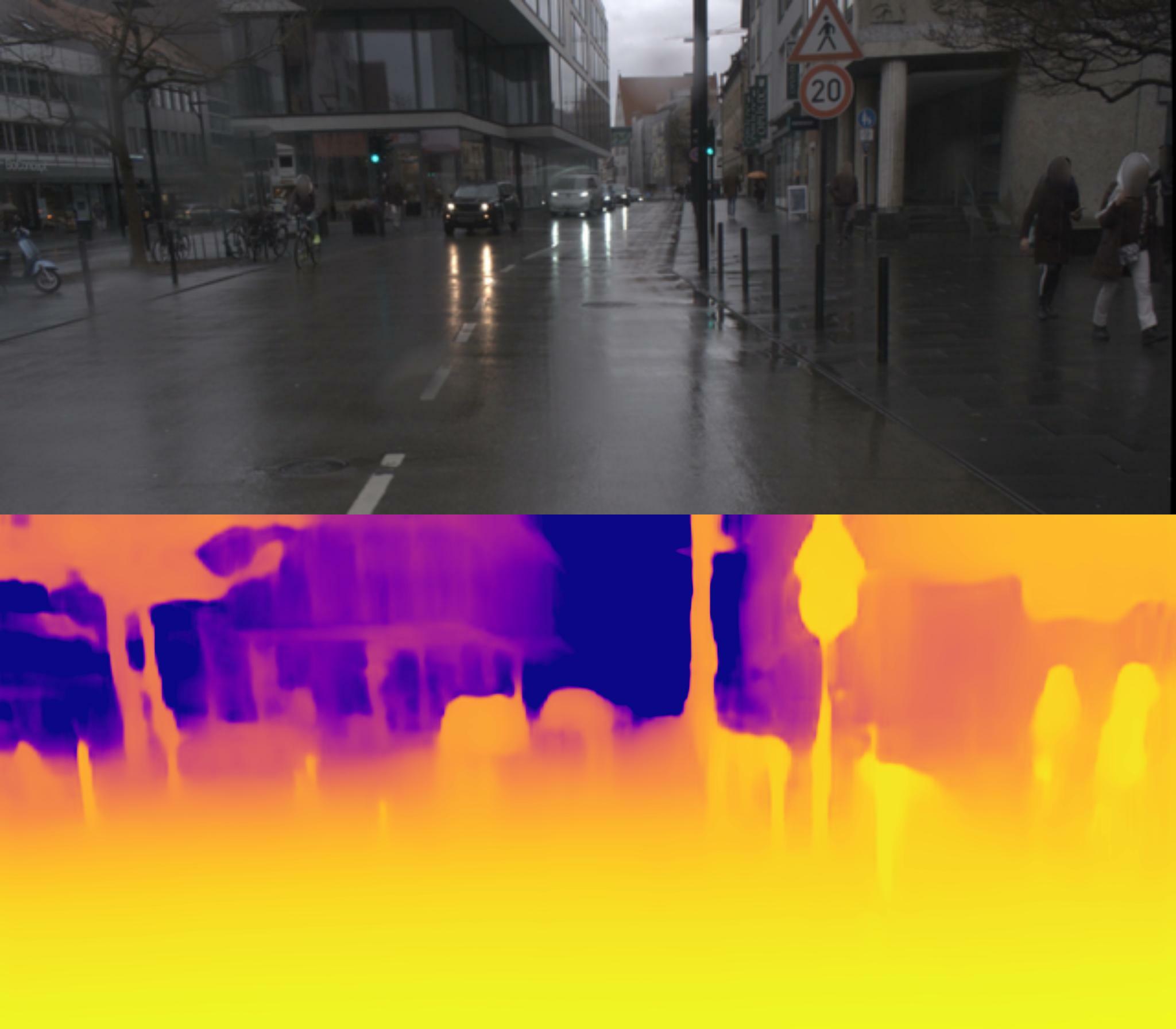}%
      \label{fig:qualitative_comp_light}
    }
    \hfill
    \subfloat[Example from the heavy weather test split.]{%
      \noindent\includegraphics[width=0.3\textwidth]{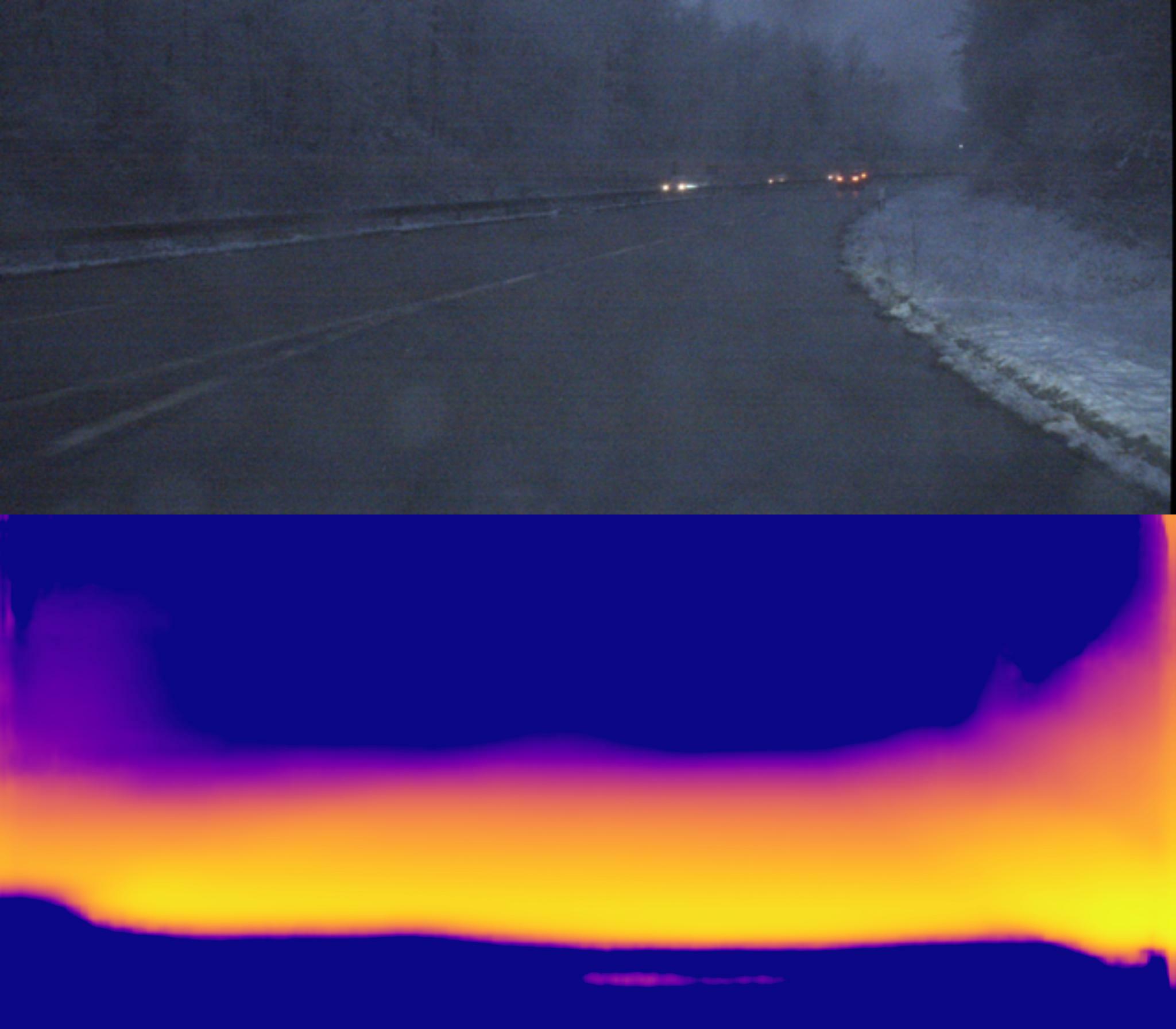}%
      \label{fig:qualitative_comp_heavy}
    }
    \caption{Depth predictions from MonoViT~\cite{zhao2022monovit} trained using monocular video sequences for different test splits of the ADUULM-360 dataset. The model generally struggles with dynamic objects moving at the same speed as the camera, as shown in (a). While light rain, as shown in (b), does not affect the results significantly, the model fails for more severe conditions such as snowfall shown in (c).}
    \label{fig:qualitative_comp}
\end{figure*}

Based on the provided experiments, we come to point out the following limitations of current state-of-the-art methods:

\textbf{State-of-the-art methods suffer under adverse weather conditions:}
As shown in Table~\ref{tab:monocular} and Table~\ref{tab:surround}, all methods share a performance drop under adverse weather conditions.
However, the drop is much higher for heavier conditions, such as heavy rain or snowfall, than for lighter adverse weather conditions.
This can also be seen when looking at qualitative examples.
Fig.~\ref{fig:qualitative_comp} shows one example of each test split for the MonoViT baseline trained using monocular video sequences.
While the depth prediction suffers only marginally under the light rain shown in Fig.~\ref{fig:qualitative_comp_light}, the prediction fails under the snowfall shown in Fig.~\ref{fig:qualitative_comp_heavy}.

\textbf{Video sequence training suffers from dynamic objects:}
In the baseline results shown in Table~\ref{tab:monocular}, the difference between video sequence and stereo training in the good weather split is much higher than in adverse weather splits.
This is because the good weather split contains more dynamic objects moving at the same speed as the ego vehicle.
This is a known limitation of self-supervised methods trained on video sequences since dynamic objects corrupt the photometric loss.
In such cases, the model predicts infinite depth for the dynamic object, as seen in the example in Fig.~\ref{fig:qualitative_comp_good}.
Most methods do not focus on this limitation since the KITTI test split does not contain many dynamic objects.

\textbf{Transformer-based approaches suffer less from adverse weather conditions:}
As shown in Table~\ref{tab:monocular}, transformer-based approaches such as MonoViT and Lite-Mono suffer less in adverse weather conditions than convolution-based approaches such as MonoDepth2.
This aligns with recent research on the robustness of depth estimation methods~\cite{kong2023robodepth}, which states that self-attention generally leads to more stable depth predictions.
However, this needs further investigation, given examples such as the failed prediction in Fig.~\ref{fig:qualitative_comp_heavy}.

\textbf{Training with adverse training data does not necessarily improve adverse weather performance:}
Section~\ref{sec:ablation} showed that training a model using good and adverse weather data does not necessarily improve performance in adverse weather conditions.
While training with good and adverse weather data could slightly improve results, fine-tuning on adverse data decreased performance under adverse weather conditions.
This also aligns with recent research~\cite{kong2023robodepth}, suggesting that more sophisticated methods are necessary to incorporate adverse weather data during training.

\textbf{Adverse weather performance depends on camera mounting position:}
Table~\ref{tab:surround} showed that the average performance decrease under adverse weather conditions is more pronounced for surround depth estimation than monocular training.
When looking at the individual cameras, the front center and right cameras show the most decreased performance.
In contrast, the rear-facing cameras suffer less under adverse weather conditions.
This can be primarily addressed by the fact that the front-facing cameras are mounted in the driving direction and thus get more disturbed by,~\eg, raindrops on the lens, compared to rear-mounted cameras.

\section{Conclusion}
This paper introduces the ADUULM-360 dataset, a novel multi-modal dataset for depth estimation in adverse weather conditions.
The dataset contains diverse scenes in different good and adverse weather conditions captured by a test vehicle with a modern sensor setup, including multiple cameras, lidar, and radar sensors.
We define several self-supervised depth estimation tasks on this dataset and provide baseline results for each task.
Our dataset provides three distinct test splits with increasing severity of adverse weather effects.
Our experiments point out shortcomings of current state-of-the-art methods and open research questions for future work.

\bibliographystyle{IEEEtran}
\bibliography{IEEEabrv,library}

\end{document}